\documentclass{article}
\usepackage{spconf,graphicx}
\usepackage{booktabs}       
\usepackage{url}
\usepackage{multicol}
\usepackage{xcolor}
\usepackage{xspace}
\usepackage{multirow}
\usepackage{comment}
\usepackage{amssymb, amsmath}
\usepackage[subrefformat=parens]{subcaption}
\usepackage{cite}
\usepackage{cleveref}

\crefname{figure}{Fig.}{Figs.}
\crefname{table}{Tab.}{Tabs.}
\crefname{equation}{Eq.}{Eqs.}
\usepackage{tabu}
\usepackage{pifont}
\newcommand{\cmark}{\ding{51}}%
\newcommand{\xmark}{\ding{55}}%
\newcommand{\ie}{i.e.\xspace}

\newcommand{\etal}{et al.\xspace}

\title{Skeleton-based Zero-Shot Spatio-Temporal Action Localization via Weakly-Supervised Pretraining}
\name{}
\address{}
\name{Koshiro Nagano$^{1,2}$$^*$~~~~~Fumiaki Sato$^3$$^\dag$$^*$~~~~~Ryo Hachiuma$^{4}$$^\dag$~~~~~Kazuki Tsutsukawa$^1$~~~~~Taiki Sekii$^{3}$$^\dag$$^\ddag$}
\address{$^1$Konica Minolta, Inc.~~~~~$^2$Keio University~~~~~$^3$CyberAgent~~~~~$^4$NVIDIA}

\begin{document}
%
\maketitle
\renewcommand{\thefootnote}{\fnsymbol{footnote}}
\footnotetext[1]{~Equal contribution.}
\footnotetext[2]{~The work was done at Konica Minolta, Inc.}
\footnotetext[3]{~Corresponding author.}
\renewcommand{\thefootnote}{\arabic{footnote}}
\setcounter{footnote}{0}

\begin{abstract}
  We propose a novel pretraining strategy for skeleton-based zero-shot spatio-temporal action localization to estimate unseen actions for person instances while overcoming high annotation costs for training via new target actions and pretraining using large-scale action scenery datasets.
  Specifically, our approach, termed {\rm Skeleton-Language feature Pooling Switching}, introduces a weakly-supervised vision-language pretraining mechanism.
  This mechanism transitions pooling kernels from pretraining, which aggregates skeleton features at the video level and aligns them with each video's known action text embeddings, to the inference phase that computes instance-level features without training via target actions.
  Furthermore, we propose {\rm Scene-Mixed Discriminative Contrastive Learning} to distinguish actions at the instance level within the combined scene through the MIL framework.
  Our experiments on four spatio-temporal action localization and classification datasets demonstrate that the proposed method effectively addresses annotation limitations.
\end{abstract}
\begin{figure*}[ht]
  \begin{center}
      \includegraphics[width=\linewidth]{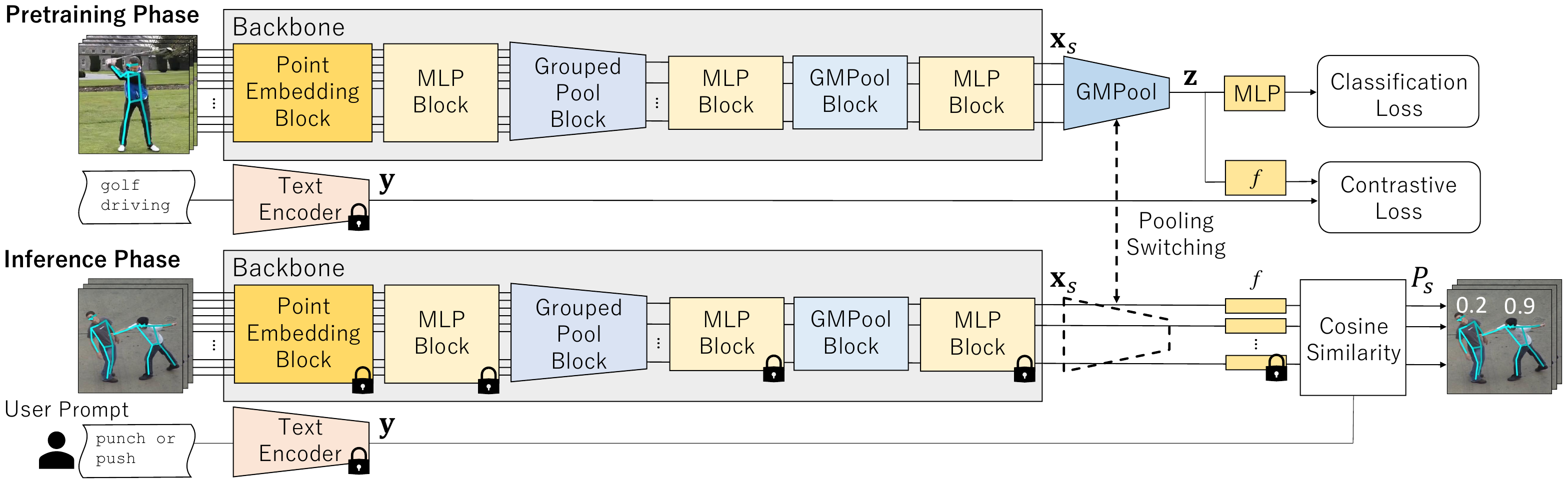}
      \caption{Overview of proposed SLPS learning framework.}
      \label{fig:overview}
  \end{center}
\end{figure*}

\section{Introduction}
\label{sec:intro}

Recognizing a person's actions in a video is crucial for various applications such as robotics~\cite{Rodomagoulakis2016ICASSP,Lee2019IROS} and surveillance cameras~\cite{Su2020ECCV,Cheng2021ICPR,Islam2021IJCNN}.
Three primary tasks have been extensively investigated in action recognition: action classification\footnote{In the conventional research, action classification is often referred to as action recognition.}, temporal action localization, and spatio-temporal action localization.
In action classification, action labels are assigned to an entire video.
In temporal action localization, action labels are assigned to each frame, and in spatio-temporal action localization, action labels are assigned to person \textit{instances}, such as the bounding boxes or skeletons detected in each video frame.
Two primary approaches have been the main focus of the study.
The first approach leverages the appearance information extracted from videos~\cite{Cheng2021ICPR,Piergiovanni_2023_CVPR}, and the second approach relies exclusively on human skeletons\footnote{Joints or keypoints specific to a person are referred to as skeletons for clarity, although some are not actual human joints.} detected in videos~\cite{Sijie2018AAAI,Duan_2022_CVPR,Su2020ECCV} to serve as inputs for deep neural networks (DNNs).
The appearance-based approaches use appearance features obtained by applying DNNs to an input video.
By contrast, the skeleton-based approaches use only low-information keypoints detected using the multi-person pose estimation methods~\cite{Sekii2018ECCV}, making these approaches relatively robust to such appearance changes in a scene or a person~\cite{Weinzaepfel2021IJCV,kas2025foundation}.
Moreover, even recent multi-modal LLMs~\cite{team2023gemini}, which heavily rely on appearance cues from video inputs, often struggle to accurately capture subtle human actions or states~\cite{salehi2024actionatlasvideoqabenchmarkdomainspecialized}, thereby revealing their limited robustness in real-world scenarios ({\it cf.} Sec. \ref{sec:exp_VLM}).
Therefore, this study aims to robustly recognize each individual's actions across diverse scenes by focusing on skeleton-based spatio-temporal action localization.

In conventional approaches, spatio-temporal action localization can be accomplished by feeding a cropped video clip~\cite{Piergiovanni_2023_CVPR} or a sequence of skeletons~\cite{Sijie2018AAAI,Duan_2022_CVPR,Su2020ECCV} into traditional action classification methods at predefined-frame intervals.
This approach, called tracking-based, is contingent upon the precise processing of person instance detections using multi-person tracking methods.
However, these tracking-based approaches have not learned the transition points of actions to accurately identify the start and end times of actions.
Conversely, the supervised approaches~\cite{Wang_2023_CVPR,Zhao_2022_CVPR} achieve precise localization by employing action labels annotated for each individual in every frame during the DNN training.
However, this method necessitates dense, instance-level annotations throughout the training process.
To mitigate the burden of annotation costs, weakly-supervised methods have been proposed.
These methods rely on a single label for the entire video for supervision~\cite{Anurag2020ECCV,Cheron2018Neurips}.
The fully or weakly-supervised approaches encounter a significant challenge when there is a need to recognize new actions not included in the training dataset (referred to as {\it target} actions).
This entails considerable human effort in collecting and annotating a vast number of video images to train the DNNs.

Furthermore, recent developments in computer vision have introduced a vision-language pretraining paradigm~\cite{pmlr-v139-radford21a} utilizing contrastive learning between image features and text embeddings.
This approach notably enhances the capability to recognize unknown object classes.
Action recognition fields have recently seen an increasing application of such a pretraining approach.
Several methods have been proposed to identify actions at the video~\cite{Rasheed_2023_CVPR,Lin_2023_ICCV,tevet2022motionclip}, frame~\cite{nag2022zero}, or instance level~\cite{Huang_2023_ICCV} in a zero-shot manner without training on the target actions.
These methods eliminate the need for the annotation costs that are typically required for the target actions.
However, the conventional zero-shot spatio-temporal action localization method~\cite{Huang_2023_ICCV, bao2025exploiting} necessitates considerable human effort in the pretraining phase.
The reason is that this method requires the annotation of action labels for each individual in every frame of a large number of pretraining videos, including actions other than the target actions, to acquire a generalized feature representation of diverse actions for DNNs.

\subsection{Overview and Contributions}\label{sec:intro_overview}
To reduce the annotation costs required for training new target actions and instance-level pretraining, we propose a novel approach to pretrain\footnote{\textit{Pretraining} refers to learning data that do not include the target actions and differs from \textit{training} that learns them.} DNNs in a weakly-supervised manner using only video-level action labels.
This approach facilitates the zero-shot detection of instances and their unknown target actions in each frame.
To achieve the weakly-supervised pretraining manner, we employ vision-language pretraining for action recognition at the video level and introduce a novel mechanism called \textit{Skeleton-Language feature Pooling Switching} (SLPS) mechanism.
It switches the pooling kernels from the pretraining, which aggregates skeleton features at the video level, to the inference phase to recognize instance-level actions.
The proposed DNN architecture incorporates a permutation-invariant structure~\cite{Qi2017CVPR} for pooling features per video in the pretraining phase and a permutation-equivariant structure~\cite{NIPS2017_f22e4747} for extracting features for each instance in the inference phase.
We apply contrastive learning to ensure that these features align with the text embeddings of action labels in a common space.
Consequently, the proposed architecture eliminates the need for multi-person tracking steps typically required in conventional methods.

Additionally, the proposed weakly-supervised pretraining approach learns to classify actions performed by individuals within a video into a single class regardless of the number of people present, based on the Multiple Instance Learning (MIL) framework~\cite{Thomas1997AI}.
However, this assumption does not hold in complex scenes with multiple people.
To mitigate this issue during pretraining, we propose a novel method called Scene-Mixed Discriminative Contrastive Learning (SM-DCL), which combines multiple scenes for pretraining.
SM-DCL merges instances from different scenes into a unified context, enabling contrastive learning to differentiate actions across instances within the combined scene.

We performed experiments to verify the effectiveness of the proposed method in terms of the aforementioned limitations of annotation costs on four datasets for spatio-temporal action localization and classification.

The main contributions of this study are summarized as follows.
(1) We define a novel task---zero-shot spatio-temporal action localization in a weakly-supervised pretraining manner---that requires no instance-level action labels in the pretraining phase.
(2) Using the proposed SLPS mechanism, we demonstrate the feasibility of such a weakly-supervised pretraining that identifies actions for each instance in the inference phase without requiring instance-level annotations over the training phase and multi-person tracking methods.
(3) We propose SM-DCL to distinguish actions at the instance level within the MIL framework.

\section{Related Work}\label{sec:related_work}
\subsection{Spatio-temporal Action Localization}
Spatio-temporal action localization is categorized into fully-~\cite{Singh_2018_ACCV,Singh_2017_ICCV,Wang_2023_CVPR} and weakly-supervised approaches~\cite{Anurag2020ECCV,Cheron2018Neurips,hfdt}.
While supervised approaches achieve high precision, they demand costly frame-wise instance annotations.
To mitigate this, weakly-supervised approaches have been proposed to utilize video-level labels.
Tracking-based methods, which identify actions in an individual skeleton sequence generated using multi-person tracking methods, have also been proposed.
However, these approaches require significant human costs for collecting many videos and annotating the target action labels when the target actions are not included in the training data.

\subsection{Zero-shot Action Recognition}
The field of vision and language has attracted considerable attention from researchers for the zero-shot visual recognition task~\cite{pmlr-v139-radford21a,Changpinyo_2021_CVPR,Cascante-Bonilla_2022_CVPR, Gupta_2022_CVPR} identifying the unseen target in the visual data with a text prompt that describes the target.
The performance can be enhanced by introducing contrastive learning~\cite{pmlr-v139-radford21a} between image features and text embeddings extracted from the prompts.
Recently, the context of vision and language has also been introduced to action recognition~\cite{Rasheed_2023_CVPR,Lin_2023_ICCV,tevet2022motionclip ,nag2022zero,ju2022prompting, Huang_2023_ICCV, bao2025exploiting, huang2025spatio}.
In spatio-temporal action localization, the appearance-based approach has been proposed~\cite{Huang_2023_ICCV, bao2025exploiting, huang2025spatio}.
This approach requires a high human effort to prepare numerous videos that include various actions and annotate instance-level labels in the pretraining phase.
Similar to the tracking-based approaches, zero-shot localization can be achieved using the information regarding individual person sequences~\cite{Rasheed_2023_CVPR,Lin_2023_ICCV,tevet2022motionclip} as inputs to zero-shot action classification methods.
Our method achieves zero-shot localization using only video-level supervision for pretraining, without relying on trackers or instance-level annotations.

\subsection{Data Mixing Augmentation}
The field of image recognition has been actively researching the data mixing augmentation~\cite{zhang2018mixup,Tokozume_2018_CVPR,yun2019cutmix,verma2019manifold,mangla2020chart,xu2020adversarial,kim2020mixco,kang2023guidedmixup,Islam_2024_CVPR}, which improves the robustness of the model by employing multiple images for data augmentation.
For instance, CutMix~\cite{yun2019cutmix} proposed to overlay a cropped area of an input image on another image to augment the data.
This approach has been applied to the weakly-supervised spatio-temporal action localization task~\cite{hachiuma2023cvpr}.
To further differentiate actions across instances in the context of MIL framework, we apply contrastive learning to the combined scenes.

\section{Proposed Method}\label{sec:method}
\cref{fig:overview} shows the overview of the proposed framework.
The DNNs in the proposed SLPS mechanism extract skeleton features for each skeletal instance (hereafter simply referred to as an instance) in the inference and pretraining phases.
In the pretraining phase, the SLPS mechanism aggregates these features into a video-level feature using Global Max-Pooling (GMPool), and based on CLIP~\cite{pmlr-v139-radford21a}, it performs contrastive learning between the video-level features and text embeddings extracted from action label names using a text encoder, such as MPNet~\cite{NEURIPS2020_c3a690be}.
Such action labels can be found in large-scale action classification datasets like Kinetics-400~\cite{Carreira2017CVPR}.
During inference, because each instance's features are aligned with the text space, the target action score is computed by measuring the similarity between the instance feature and the text embedding of the user-provided prompt, following the CLIP framework.

\subsection{Inference Phase}\label{sec:inference}
In the pretraining and inference phases, the multi-person pose estimation~\cite{Sekii2018ECCV} is applied to the input video to extract $FSK$ skeleton keypoints, where $F$, $S$, and $K$ denote the number of frames in the video clip, the number of skeletal instances per frame, and the number of keypoints per instance, respectively.
Each keypoint is then transformed into an input vector $\mathbf{v}$ for the DNNs.
The input is a five-dimensional vector consisting of the two-dimensional keypoint coordinates on the image, time index, keypoint confidence, and keypoint index.
If the number of detected skeletons or joints in each frame is fewer than $S$ or $K$, respectively, the input tensor is padded with zero vectors.

In the inference phase, the feature vector $\mathbf{x}_s$ for each instance $s \in \{1,\ldots,FS\}$, aligned in a common space with the text embeddings described in \cref{sec:pretraining}, is obtained by inputting the above input vectors into the architecture described in \cref{sec:vlps}.
Consequently, given $T$ text embeddings $\mathcal{Y}=\{\mathbf{y}_1, \ldots ,\mathbf{y}_T\}$ extracted by a text encoder for each target action and the feature vector $\mathbf{x}_s$ for each instance $s$, the score $P_s$, representing each instance $s$ includes the target action specified by the user, is formulated as follows:
\begin{equation}\label{eq:PAS}
    \begin{aligned}
    P_s = \mathrm{max}\left(\mathrm{Cos}\left(f(\mathbf{x}_s), \mathbf{y}_1\right), \ldots , \mathrm{Cos}\left(f(\mathbf{x}_s), \mathbf{y}_T\right)\right),
    \end{aligned}
\end{equation}
where $\mathrm{Cos}(\cdot,\cdot)$ represents the cosine similarity between two vectors, and $f$ denotes pretrained multilayer perceptron (MLP) to align the dimension of $\mathbf{x}_s$ and $\mathbf{y}$.

\subsection{Pretraining Phase}\label{sec:pretraining}
In the pretraining phase, the DNNs in the proposed SLPS mechanism extract the features $\mathbf{x}_s$ for each instance $s$ similar to the inference phase.
In contrast to the inference phase, GMPool aggregates these features into a single video-level feature $\mathbf{z}$ representing the entire video.
This enables the pretraining scheme to introduce a weakly-supervised learning framework based on MIL~\cite{Thomas1997AI}.

We use contrastive learning between the video-level skeleton features and the text embeddings extracted from action label names and multitask learning on the action classification task.
To achieve the MIL-based pretraining that uses the video-level action labels, we define the total loss $\mathcal{L}$ consisting of the action classification loss $\mathcal{L}_\text{cls}$ and the contrastive loss $\mathcal{L}_\text{cont}$ in a batch of $B$ videos as follows:
\begin{equation}
  \mathcal{L} = \alpha \sum_{i=1}^{B} \mathcal{L}_{\text{cls},i} + (1 - \alpha) \mathcal{L}_\text{cont},
\end{equation}
where $\alpha$ is the mixing ratio of the loss functions.
The classification loss $\mathcal{L}_{\text{cls}}$ uses the cross-entropy loss 
over $C$ action classes, where the logits are computed by feeding the video-level skeleton feature, aggregated by GMPool, into a fully-connected classifier.

Based on the loss function proposed by CLIP~\cite{pmlr-v139-radford21a}, the contrastive loss is $\mathcal{L}_\text{cont} = (\mathcal{L}_\text{s2t} + \mathcal{L}_\text{t2s})/2$.
We use a projection head $f(\cdot)$ to map $\mathbf{z}$ to the embedding space.
Here, $\mathcal{L}_\text{s2t}$ contrasts each projected video feature $f(\mathbf{z}_i)$ with all text embeddings $\{\mathbf{y}_j\}_{j=1}^{B}$ in the batch:
\begin{equation} \mathcal{L}_\text{s2t} = - \sum_{i=1}^{B} \log \frac{\exp\left(\mathrm{Cos}\left(f\left(\mathbf{z}_i\right), \mathbf{y}_i\right)/\tau\right)}{\sum_{j=1}^{B}\exp\left(\mathrm{Cos}\left(f\left(\mathbf{z}_i\right), \mathbf{y}_j\right)/\tau\right)}, 
\end{equation}
where $(\mathbf{z}_i,\mathbf{y}_i)$ is a positive pair and $\tau$ is a learnable temperature parameter.
$\mathcal{L}_{\text{t2s}}$ is defined symmetrically by swapping the roles of video and text.

\subsection{SLPS DNN Architecture}\label{sec:vlps}
\noindent \textbf{Overview.} 
For weakly-supervised pretraining, the DNN architecture of the proposed SLPS mechanism (\cref{fig:overview}) incorporates a Pooling-Switching mechanism.
This mechanism employs GMPool to transition the unit of feature aggregation from the pretraining phase to the inference phase.
In the pretraining and inference phases, the feature vectors for each instance are computed using a common set of multiple blocks.
The Point Embedding block embeds the input per keypoint vector into a high-dimensional feature vector using MLPs.
The Grouped Pool Block, described in \cref{sec:gpb}, aggregates the feature vectors transformed in the MLP block, considering the relationships between keypoints into a single feature vector for each group belonging to the same instance.
Subsequently, the MLP and GMPool blocks compute the feature vectors for each instance by considering the relationships between instances.
In the inference phase, the features for each instance are used to calculate scores using \cref{eq:PAS} as described in \cref{sec:inference}.
Meanwhile, in the pretraining phase, GMPool aggregates the features into a video-level feature, which is subsequently aligned with the text embedding space of action labels, as described in \cref{sec:pretraining}.

Notably, Max-Pooling only selects the features without requiring any transformation.
This ensures that in the pretraining phase, the aggregated video-level feature, aligned with the text embedding space, is aligned with the same space as instance-level features just before being aggregated by GMPool.
Each block is designed based on the keypoint pooling method~\cite{hachiuma2023cvpr}, and for detailed information on the MLP and GMPool blocks, please refer to \cite{hachiuma2023cvpr}.

\subsection{SM-DCL}\label{sec:sm-dcl}

\begin{figure}[tb]
  \centering
  \includegraphics[width=\hsize]{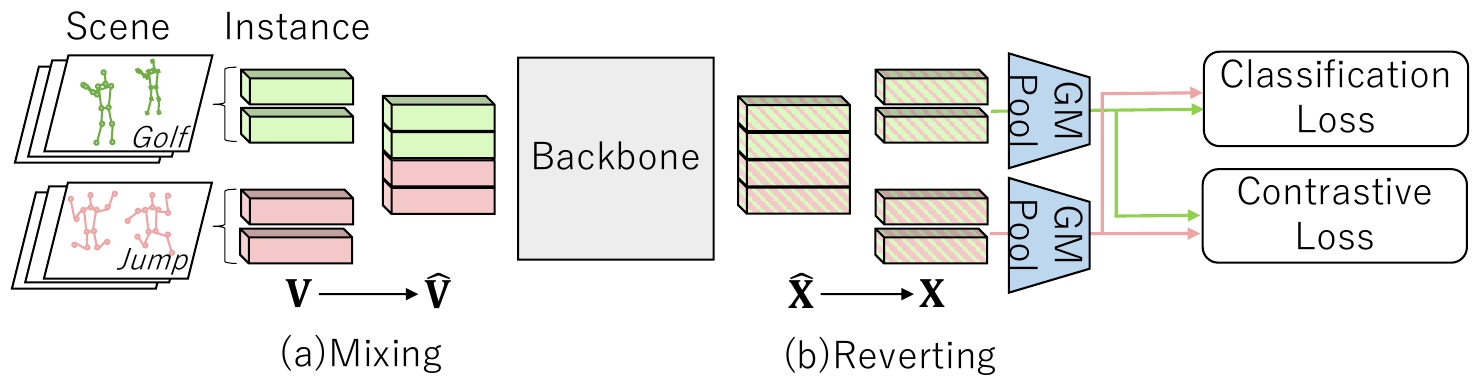}
  \caption{Overview of Scene-Mixed Discriminative Contrastive Learning for zero-shot spatio-temporal action localization. The modules same as those in \cref{fig:overview} are abbreviated with the same color.}
  \label{fig:sm-dcl}
\end{figure}

To mitigate the pretraining assumption that each video contains only a single action class even when multiple people interact, we propose a novel contrastive learning method called SM-DCL (\cref{fig:sm-dcl}). 
This method artificially mixes scenes to ensure features remain discriminative across instances.
SM-DCL employs two tensor operations: (a) reshaping the input $\mathbf{V} \in \mathbb{R}^{B \times FSK \times 5}$ to $\hat{\mathbf{V}} \in \mathbb{R}^{B/m \times mFSK \times 5}$ to overlay $m$ scenes (visualized as $m=2$ in \cref{fig:sm-dcl}), and (b) reverting the backbone output $\hat{\mathbf{X}}$ back to the unmixed state $\mathbf{X} \in \mathbb{R}^{B \times FS \times D}$ before pooling.
This allows the loss calculation to remain identical to \cref{sec:pretraining} while enhancing instance-level discriminability with minimal computational overhead.
\section{EXPERIMENTS}
\begin{table*}[tb]
\centering
\noindent

\begin{minipage}[t]{\textwidth}
\centering
\caption{Performance comparison with conventional methods.
(a) spatio-temporal action localization methods on UCF101-24,
(b) tracking-based methods on FDD,
and (c) skeleton-based video classification methods on RWF and MF.
Conventional methods are trained in a supervised (\dag) or weakly-supervised manner (\ddag).
}
\label{tab:combined_results}

\begin{minipage}[t]{\linewidth}
  \begin{minipage}[t]{0.32\linewidth}
    \vspace{0pt}\centering
    \subcaption*{\textbf{(a) UCF101-24}}
    \scalebox{0.8}{
      \begin{tabular}{c|c|cc} \hline
        \multirow{4}{*}{Method} & \multirow{4}{*}{AP (\%)} &
        \multirow{4}{*}{\shortstack{Learn \\target act. \\ in train. (\xmark) \\ or not (\cmark)}} \\
        & & \\ & & \\ & & \\ \hline
        Ch\'{e}ron~et al.~\cite{Cheron2018Neurips} \ddag & $17.7$ & \multirow{2}{*}{\xmark} \\
        Anurag~et al.~\cite{Anurag2020ECCV} \ddag & $35.0$ & \\ \hline
        Ours & $34.1$ & \cmark \\ \hline
      \end{tabular}
    }
  \end{minipage}\hspace{1mm}%
  \begin{minipage}[t]{0.32\linewidth}
    \vspace{0pt}\centering
    \subcaption*{\textbf{(b) FDD}}
    \scalebox{0.8}{
      \begin{tabular}{c|c|ccccc} \hline
        \multirow{4}{*}{Method} & \multirow{4}{*}{\shortstack{Frame \\ AP (\%)}} &
        \multirow{4}{*}{\shortstack{Learn \\ target act. \\ in train. (\xmark) \\ or not (\cmark)}} \\
        & & \\ & & \\ & & \\ \hline
        HFDT \dag & $61.6$ & \multirow{3}{*}{\xmark} \\
        CTR-GCN~\cite{chen2021channel} \dag & $36.0$ & \\
        ST-GCN++~\cite{duan2022pyskl} \dag & $45.0$ & \\ \hline
        Ours & $73.3$ & \cmark \\ \hline
      \end{tabular}
    }
  \end{minipage}\hspace{1mm}%
  \begin{minipage}[t]{0.32\linewidth}
    \vspace{0pt}\centering
    \subcaption*{\textbf{(c) RWF \& MF}}
    \scalebox{0.8}{
      \begin{tabular}{c|cc|cc} \hline
        \multirow{4}{*}{Method} & \multirow{4}{*}{RWF} & \multirow{4}{*}{MF} &
        \multirow{4}{*}{\shortstack{Learn \\ target act. \\ in train. (\xmark) \\ or not (\cmark)}} \\
        & & & \\ & & & \\ & & & \\ \hline
        PointNet++~\cite{Su2020ECCV} \dag & $78.2$ & $89.2$ &  \\
        DGCNN~\cite{Su2020ECCV} \dag & $80.6$ & $91.3$ & \xmark  \\
        SPIL~\cite{Su2020ECCV} \dag & $89.3$ & $98.5$ &  \\ \hline
        Ours & $84.0$ & $92.7$ & \cmark \\ \hline
      \end{tabular}
    }
  \end{minipage}
\end{minipage}

\end{minipage}%
\end{table*}

\begin{table}[t]
\centering
\captionof{table}{Performance comparison for ablation studies on UCF101-24 ({\it cf.} Sec.~\ref{sec:exp_ablation}).}
\label{tab:few_learning}

\scalebox{0.9}{
  \begin{tabular}{c|c|c} \hline
    Architecture & Training strategy & AP (\%) \\ \hline
    \multirow{5}{*}{Ours} & 3-shot & $33.9$ \\
                          & 2-shot & $29.9$ \\
                          & 1-shot & $23.4$ \\ \cline{2-3}
                          & w/o SM-DCL & $31.2$ \\
                          & w/ SM-DCL & $\mathbf{34.1}$ \\ 
    \hline
  \end{tabular}
}
\end{table}

\begin{table}[t]
\centering
\captionof{table}{Comparison with appearance-based methods on RWF-2000.
}
\label{tab:appearance_rwf2000}
\scalebox{0.9}{
  \begin{tabular}{l|c} \hline
    Method & Acc. (\%) \\ \hline
    X-CLIP~\cite{XCLIP} & $69.3$ \\
    EVA~\cite{Fang_2023_CVPR} & $71.3$ \\
    ViFi-CLIP~\cite{Rasheed_2023_CVPR} & $72.8$ \\
    Qwen2.5-VL-7B-Instruct~\cite{bai2025qwen25vl} & $80.0$ \\ \hline
    \textbf{Ours} & $\mathbf{84.0}$ \\ \hline
  \end{tabular}
}
\end{table}

\subsection{Datasets and Evaluation Settings}
\noindent \textbf{Kinetics-400} is a large-scale action recognition dataset collected from YouTube videos with $400$ action classes~\cite{Carreira2017CVPR}.
We utilize this dataset for pretraining the DNNs.
Among the target action classes in the subsequent datasets, Kinetics-400 does not include labels for violent and falling behaviors.

\noindent \textbf{UCF101-24} is a subset of the UCF101 dataset~\cite{Soomro2012Arxiv}, a specific action classification dataset composed of videos from YouTube.
This dataset consists of videos belonging to $24$ action classes with bounding-box level annotations.
We use this dataset to compare the proposed method with conventional weakly-supervised spatio-temporal action localization methods~\cite{Anurag2020ECCV,Cheron2018Neurips}.
In line with previous studies of zero-shot action classification~\cite{Rasheed_2023_CVPR,Lin_2023_ICCV}, the experiment excludes overlapped five action classes to compare performance using Kinetics-400 for pretraining.

\noindent \textbf{FDD}
is a spatio-temporal action localization dataset designed by the authors for comparing a SoTA conventional tracking-based approach (HFDT\footnote{\scriptsize\url{https://github.com/GajuuzZ/Human-Falling-Detect-Tracks}}) that utilizes ST-GCN~\cite{Sijie2018AAAI} with the proposed method.
Since the supervised fall-action dataset Le2i~\cite{Le2i2013JEI}, used to train HFDT, is not publicly available, we captured videos under shooting conditions similar to Le2i.
We employ frame average precision (Frame AP) as the evaluation metric for spatio-temporal action localization, followed by the AVA dataset~\cite{Gu_2018_CVPR}.
Among the seven actions, we categorize all labels related to falling behavior collectively as falls.

\noindent \textbf{RWF-2000 and MF}
are violent action classification datasets.
We evaluate the proposed method on these datasets by comparing it with supervised baselines.
For fairness, we classify each video using the highest instance score for the violent-action prompt.

For additional details on the implementation and dataset, please refer to the Supplementary Material.

\subsection{Comparative Experiments on Conventional Methods}\label{sec:comp_exp}
\cref{tab:combined_results} (a) to (c) summarize the performances of the proposed method and the conventional fully and weakly-supervised methods.

As shown in \cref{tab:combined_results} (a), the proposed method outperforms the conventional weakly-supervised method~\cite{Cheron2018Neurips} in terms of spatio-temporal action localization accuracy although UCF101-24 is an advantageous dataset for appearance-based approaches~\cite{hachiuma2023cvpr}.

As shown in \cref{tab:combined_results} (b), the proposed method considerably outperforms conventional tracking-based methods~\cite{hfdt, chen2021channel, duan2022pyskl} in terms of spatio-temporal action localization accuracy.
The conventional methods shown in Tabs. \ref{tab:combined_results}(b) and (c) use more accurate human pose detectors, AlphaPose ($72.0$~$\text{\%AP}_\text{kp}$), RMPE~\cite{Fang_2017_ICCV} ($72.3$~$\text{\%AP}_\text{kp}$), and HRNet ($74.6$~$\text{\%AP}_\text{kp}$), compared with the PPNs ($60.6$~$\text{\%AP}_\text{kp}$) employed in the proposed method.
\footnote{The result indicates the accuracy of the human pose detection on the MS-COCO validation set.}
The discrepancy in the accuracy of human pose detection shows that the tracking-based method yields poor spatio-temporal localization accuracy (Sec. \ref{sec:intro}).

\begin{figure}[ht]
\centering
\includegraphics[width=0.75\linewidth]{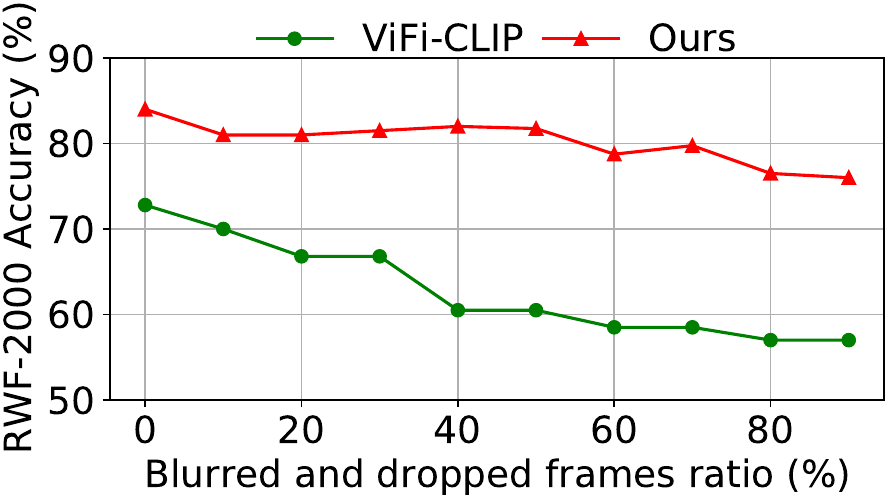}
\captionof{figure}{Robustness against degradation of input videos.}
\label{fig:noise}
\end{figure}
As shown in \cref{tab:combined_results} (c), the proposed method outperforms many conventional supervised methods, including PointNet++~\cite{Su2020ECCV} and DGCNN~\cite{Su2020ECCV}, in terms of violence action classification accuracy.
Its accuracy is also comparable to that of SPIL~\cite{Su2020ECCV}.
This enhanced accuracy is achieved despite leveraging the outputs of our spatio-temporal action localization method, which differs fundamentally from action classification.
These results indicate that the proposed method attains an accuracy comparable with existing supervised approaches that employ SoTA architectures for action classification tasks.

In summary, the proposed method successfully achieves zero-shot spatio-temporal action localization with a certain degree of accuracy without target action-specific DNN training and by exclusively relying on video-level annotations as supervision during pretraining.
Consequently, it effectively reduces the annotation costs required for pretraining and training new target actions (Sec. \ref{sec:intro}).

\subsection{Ablation Studies}\label{sec:exp_ablation}
\noindent \textbf{In-depth Analysis of Individual Components.}
\cref{tab:few_learning} shows the results of quantitatively validating each contribution described in \cref{sec:intro} using the UCF101-24.
Rows 2--4 of \cref{tab:few_learning} (denoted as {\it $n$-shot} in the training strategy) show a comparison of the spatio-temporal action localization accuracy of models trained with few-shot learning using instance-level labels.
{\it $n$-shot} corresponds to learning from randomly selected $n$ videos per action.
Each model is fine-tuned after pretraining on the Kinetics-400 dataset, similar to the proposed method, to ensure fair comparison.
The average accuracy is reported for three models evaluated using different few-shot subsets.
The accuracy in the three-shot setting ($33.9$\%AP) is comparable with that of the proposed method ($34.1$\%AP in \textit{Ours w/ SM-DCL}), indicating that weakly-supervised pretraining reduces the annotation cost of manual three-video labeling per action (over ten hours).

\noindent \textbf{SM-DCL.}
Rows 5--6 of \cref{tab:few_learning} (denoted as {\it Ours w/} or {\it w/o SM-DCL} in the training strategy) show a comparison of the accuracy of models trained with or without SM-DCL.
Notably, the proposed method with SM-DCL outperforms that without SM-DCL ($34.1$\%AP vs. $31.2$\%AP), indicating that the process of mixing scenes enhances the discriminability of actions for each instance during pretraining.


\noindent \textbf{Comparison with video foundation models.}\label{sec:exp_VLM}
\cref{tab:appearance_rwf2000} compares appearance-based zero-shot methods and representative large vision-language models (LVLMs) with ours on RWF-2000.
\cref{fig:noise} shows how the performance of the appearance‐based method~\cite{Rasheed_2023_CVPR} and proposed method degrades with decreasing input image quality owing to blur and mask-induced frame drops on this dataset.
Because of the diversity of scenes, varying appearances of individuals, and degradation of the input video, the proposed method exhibits higher robustness than appearance‐based approaches.

While recent LVLMs typically contain more than 7B parameters and are pretrained on massive Internet-scale corpora, our model contains only about 70M parameters, approximately 100$\times$ fewer, and is trained using only Kinetics-400.
On RWF-2000, Qwen2.5-VL-7B-Instruct achieves 80\% accuracy, which is lower than that of our method.
Moreover, Qwen2.5-VL-7B-Instruct runs at approximately 8 FPS, whereas our method achieves approximately 1900 FPS\cite{hachiuma2023cvpr}, corresponding to about a 240$\times$ speedup.
These results demonstrate that our method achieves competitive or superior recognition performance with substantially lower computational and pretraining requirements.

\section{CONCLUSION}
This paper proposed a novel skeleton-based zero-shot spatio-temporal action localization method that estimates unseen actions for person instances detected in each video frame, addressing annotation limitations in existing localization methods.
The core challenges are threefold: 
(1) Introducing a novel task---zero-shot spatio-temporal action localization in a weakly-supervised pretraining manner. This involves learning only video-level action labels during the pretraining phase without training via the target actions.
(2) Demonstrating the feasibility of such a weakly-supervised pretraining that identifies actions for each instance in the inference phase without requiring instance-level annotations over the pretraining phase and without relying on multi-person tracking methods.
(3) Proposing Scene-Mixed Discriminative Contrastive Learning (SM-DCL) to enable the distinction of actions at the instance level within the MIL framework.
In the experiments, we evaluated the effectiveness of the proposed method against annotation limitations.
\vfill\pagebreak

\bibliographystyle{IEEEbib}
\bibliography{refs}

\setcounter{section}{5}
\setcounter{figure}{3}
\setcounter{table}{3}
\setcounter{equation}{4}

\clearpage
\setcounter{page}{1}

\let\titleold\title
\renewcommand{\title}[1]{\titleold{#1}\newcommand{\thetitle}{#1}}
\def\maketitlesupplementary
   {
   \newpage
       \twocolumn[
        \centering
        \large
        \textbf{\thetitle}\\
        \vspace{0.5em}Supplementary Material \\
        \vspace{1.0em}
       ] 
   }
\title{SKELETON-BASED ZERO-SHOT SPATIO-TEMPORAL ACTION LOCALIZATION VIA WEAKLY-SUPERVISED PRETRAINING}

\maketitlesupplementary

\section{Experiments}

\subsection{Dataset}
\noindent \textbf{Kinetics-400.} 
Kinetics-400~\cite{Carreira2017CVPR} is a large-scale action recognition dataset collected from YouTube\footnote{\url{youtube.com}} videos with $400$ action classes.
It contains $250$K training and $19$K validation $10$-second video clips with $30$ fps.

\noindent \textbf{UCF101-24.}
The UCF101-24 dataset~\cite{Soomro2012Arxiv} is a subset of the UCF101 dataset, a specific action classification dataset composed of videos from YouTube.
UCF101-24 consists of videos belonging to $24$ action classes and comprises $3.2$K videos.
Additionally, its $24$ class action labels are annotated for each bounding box in the videos.
We use the corrected annotation~\cite{Singh2017ICCV} according to the standard practice~\cite{Cheron2018Neurips,Anurag2020ECCV}.

\noindent \textbf{FDD.} 
The FDD dataset is a spatio-temporal action localization dataset designed by the authors for comparing a SoTA conventional tracking-based approach (HFDT)~\cite{hfdt} that utilizes ST-GCN~\cite{Sijie2018AAAI} with the proposed method.
The supervised fall action classification dataset, Le2i~\cite{Le2i2013JEI}, is used in the training of the conventional method.
This dataset is not publicly available at the moment.
Thus, in the FDD dataset, we captured videos under shooting conditions similar to those of the Le2i dataset, including the size of the subjects, camera angle, and indoor environment, to compare localization accuracy fairly between the conventional and proposed methods.

The dataset comprises $62$ video clips, of which $33$ feature individuals simulating falls.
Each clip lasts approximately $15$ seconds and is recorded under conditions that ensure consistent visibility of $1$ to $2$ individuals in each video.
The seven action labels, which semantically subdivide falls and non-falls similar to Le2i, are annotated for each bounding box in the videos.

Note that HFDT~\cite{hfdt} is trained on the Le2i, while CTR-GCN~\cite{chen2021channel} and ST-GCN++~\cite{duan2022pyskl} are trained on NTU RGB+D 120~\cite{liu2020ntu}, a relatively large-scale action classification dataset that includes fall actions.

\noindent \textbf{RWF-2000.}
RWF-2000~\cite{Cheng2021ICPR} is the violent action classification dataset collected from YouTube videos.
The videos feature two actions, violent or non-violent, captured by security cameras with different people and backgrounds.
There are $1.6$K training and $0.4$K test $5$-second video clips with $30$ fps.
Two-class labels are annotated for each video.

\noindent \textbf{MF.}
The MF dataset~\cite{nievas2011violence} is a collection of $200$ video clips extracted from action movies, specifically designed for assessing violent action classification. 
A wider variety of fight scenes were captured at different resolutions, and nonfight videos were extracted from public action-recognition datasets.

\begin{table*}[t]
    \centering
    \begin{minipage}{0.5\linewidth}
      \caption{Hyperparameters of each dataset during pretraining.}
      \centering
      \scalebox{0.7}{
      \begin{tabular}{c|cccc} \hline 
          Evaluation dataset & \multicolumn{2}{c}{UCF101-24~\cite{Soomro2012Arxiv} and MF~\cite{nievas2011violence}} & \multicolumn{2}{c}{FDD and RWF-2000~\cite{Cheng2021ICPR}} \\ \hline
          Pretraining dataset & \multicolumn{4}{c}{Kinetics-400~\cite{Carreira2017CVPR}} \\
          Pose Detector & \multicolumn{2}{c}{HRNet~\cite{Sun2019CVPR}} & \multicolumn{2}{c}{PPNs~\cite{Sekii2018ECCV}} \\ \hline
          Mixing ratio & \multirow{2}{*}{$1$} & \multirow{2}{*}{$0.6$} & \multirow{2}{*}{$1$} & \multirow{2}{*}{$0.6$} \\
          of the loss functions $\alpha$ & & & & \\ \hline 
          Optimizer        & \multicolumn{4}{c}{Stochastic Gradient Descent}                         \\
          Number of epochs & $150$ & $40$ & $150$ & $40$ \\ 
          Batch size       & $480$ & $256$ & $480$ & $256$ \\ 
          Learning rate    & $0.48$ & $1.6$ & $0.48$ & $1.6$ \\ 
          Weight decay     & $0.00005$ & $0.0000125$ & $0.00005$ & 0.0001 \\ 
          Momentum         & \multicolumn{4}{c}{$0$}                                                       \\
          LR scheduler     & \multicolumn{4}{c}{linear}                                                     \\  
          Joint scaling & \multicolumn{4}{c}{$[0.8,1.2]$} \\
          Joint shift   & \multicolumn{4}{c}{$0.2$} \\ 
          Joint rotate $(^\circ)$ & $10$ & $5$ & $10$ & $5$ \\ 
          Joint flip ratio   & \multicolumn{4}{c}{$0.5$}                                                  \\ 
          Temporal crop window   & \multicolumn{4}{c}{$100$} \\ 
          Temporal FPS drop & \multicolumn{4}{c}{$5$} \\ \hline
      \end{tabular}
      \label{tab:hp}}
      \end{minipage}
      \begin{minipage}{0.45\linewidth}
      \centering
      \includegraphics[width=0.8\linewidth]{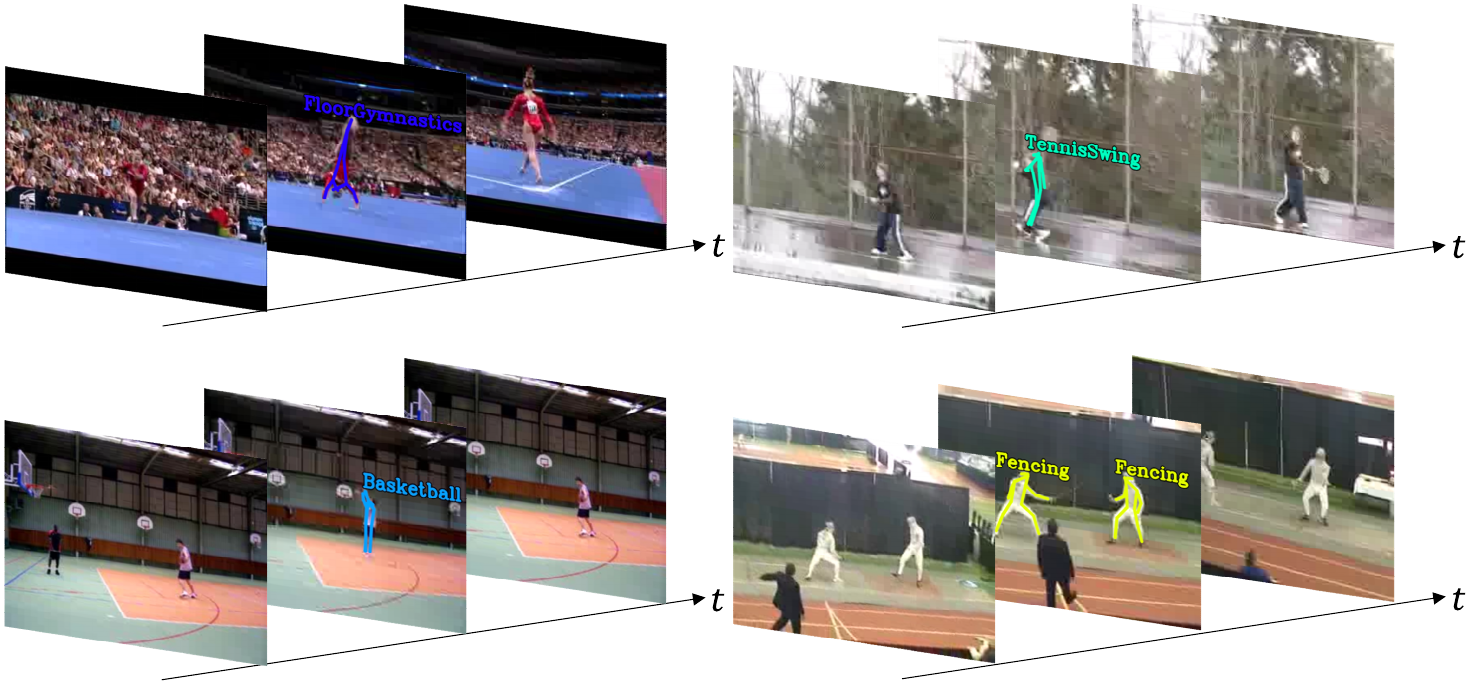}
      \captionof{figure}{Qualitative zero-shot spatio-temporal action localization results of the proposed method on UCF101-24 ({\it cf.} \cref{sec:comp_exp}). The input keypoints and the user prompts are visualized in the figure. Our approach does not require instance-level annotations during training or multi-person tracking methods, yet it can localize actions spatio-temporally.}
      \label{fig:qualitative_results} 
      \end{minipage}
\end{table*}

\subsection{Evaluation Settings}
\noindent \textbf{UCF101-24} is used to compare the proposed method with conventional weakly-supervised spatio-temporal action localization methods~\cite{Anurag2020ECCV,Cheron2018Neurips}.
Similar to conventional methods, we use video average precision (Video AP)~($\%$) with 3D IoU=$0.5$ as the evaluation metric for the spatio-temporal action localization.
We also employ the tubelet generated by Ch\'{e}ron~\etal~\cite{Cheron2018Neurips} and link the tubelets using the linking algorithm~\cite{Kalogeiton2017ICCV}, followed by the conventional method~\cite{Anurag2020ECCV}.

\noindent \textbf{FDD} is used to compare the conventional tracking-based methods~\cite{hfdt, chen2021channel, duan2022pyskl} with the proposed method.
We employ frame average precision (Frame AP) as the evaluation metric for spatio-temporal action localization, followed by the AVA dataset~\cite{Gu_2018_CVPR}.
Among the seven actions, we categorize all labels related to falling behavior collectively as falls.

\noindent \textbf{RWF-2000 and MF.}
Recently, skeleton-based methods have been actively studied in the field of action classification.
In contrast to spatio-temporal action localization, SoTA action classification architectures are subject to frequent updates.
We employ this dataset to validate the effectiveness of the proposed method by comparing its classification accuracy with that of conventional supervised action classification methods.
For a fair comparison with the conventional methods, the proposed method classifies each video as either violent or non-violent using the highest instance score for violent action prompts within the video.

\subsection{Implementation Details}
\subsubsection{Parameters of SLPS DNN Architecture}
The dimension of the output feature vector at the point embedding block is set to $256$ and those of the input feature vector at the MLP blocks are set to $256$, $512$, and $1024$ because the Grouped Pool and GMPool blocks double the channels via concatenation.
The MLP expansion ratio is set to $2.0$.
Batch normalization~\cite{Ioffe2015ICML} is employed as the normalization layer, and GELU~\cite{Hendrycks2016Arxiv} is employed as the activation function in the MLP block.

\subsubsection{Grouped pool block.} \label{sec:gpb}
The Grouped Pool Block consists of GMPool $\phi_G$ and Local Max-Pooling (LMPool) $\phi_L$, which applies Max-Pooling to each local group of the same instance to which the input feature vector per keypoint belong.
The Grouped Pool Block outputs $FS$ feature vectors containing the number of instances in the video.
The Grouped Pool Block can be expressed as follows:
\begin{equation}
  \mathbf{U}_\text{out} = \left\{\left[\phi_L\left(\mathbf{U}_{\text{in},s}\right),\phi_G\left(\mathbf{U}_\text{in}\right)\right]\right\}_{s \in \left\{1,\ldots,FS\right\}}.
  \label{eq:gpb}
\end{equation}
$\mathbf{U}_\text{in} \in \mathbb{R}^{FSK \times D}$ and $\mathbf{U}_\text{out} \in \mathbb{R}^{FS \times 2D}$ are the matrices of the input and output feature vectors, respectively, as described below.
Then, the input feature vector $\mathbf{u}_{\text{in}} \in \mathbf{U}_\text{in}$ per keypoint is grouped into $FS$ instances $\left(\mathbf{u}_{\text{in}} \in \mathbb{R}^{1\times D}\right)$, and $\mathbf{U}_\text{in}$ can be expressed by a concatenated matrix as follows:
\begin{equation}
  \mathbf{U}_\text{in} = \left(\mathbf{u}_{\text{in}, 1};\ldots;\mathbf{u}_{\text{in}, FSK}\right) = \left(\mathbf{U}_{\text{in}, 1};\ldots;\mathbf{U}_{\text{in}, FS}\right),
\end{equation}
where $K$ is the number of keypoints.
$\mathbf{U}_{\text{in}, s}$ represents a submatrix of $\mathbf{U}_\text{in}$ for each instance $s$.
Consequently, $\mathbf{U}_\text{out}$ is computed using the output vector $\mathbf{u}_{\text{out}} \in \mathbb{R}^{1\times D}$ as follows:
\begin{equation}
\mathbf{U}_\text{out} = {\left(\mathbf{u}_{\text{out}, 1};\ldots;\mathbf{u}_{\text{out}, FS}\right)}^T.
\end{equation}

In \cref{eq:gpb}, we concatenate each feature vector $\phi_L\left(\mathbf{U}_{\text{in},s}\right) \in \mathbb{R}^{1\times D}$ computed for the instance group $s$ and the global feature vector $\phi_G\left(\mathbf{U}_\text{in}\right) \in \mathbb{R}^{1\times D}$ in a channel dimension.
Moreover, LMPool~$\phi_L(\cdot)$ can be expressed as follows:
\begin{equation}
  \begin{split}
    \phi_L\left(\mathbf{U}_{\text{in},s}\right) = \mathrm{MaxPool}(\mathbf{U}_{\text{in}, s}),
  \end{split}
\end{equation}
where $\mathrm{MaxPool}(\cdot)$ is the operation used to obtain the max value for each channel from the feature vectors matrices.
$\phi_G(\cdot)$ is expressed as follows:
\begin{equation}
  \phi_G(\mathbf{U}_\text{in}) = \mathrm{MaxPool}(\mathbf{U}_\text{in}).
\end{equation}

In the pretraining phase, given the instance-level feature $(\mathbf{x}_1,\ldots ,\mathbf{x}_{FS})$, GMPool aggregates these features in the video-level feature $\mathbf{z}$ as follows:
\begin{equation}
  \mathbf{z} = \mathrm{MaxPool}\left(\mathbf{x}_1;\ldots;\mathbf{x}_{FS}\right).
  \label{eq:poolingswitching}
\end{equation}
Note that GMPool operation in \cref{eq:poolingswitching} is removed in the inference phase, based on the Pooling-Switching.

\subsubsection{Data Augmentation}
We apply two types of data augmentation during pretraining: augmentation onto the image space and along the temporal axis to the input joints. 
We then randomly scale, shift, rotate, and flip the joint coordinates for augmentation onto the image space and randomly crop the input joints with a random size of the temporal window. 
Additionally, we drop joints within a random interval range.

\subsubsection{Hyperparameters}
\cref{tab:hp} shows the hyperparameters employed in each experiment.
We use stochastic gradient descent for DNN pretraining with a learning rate that decreases with the number of epochs.
The model is pretrained for $150$ epochs on Kinetics-400 with $\alpha=1$.
After these epochs, we set $\alpha$ to $0.6$ to pretrain the model for $40$ epochs.
In SM-DCL (\cref{sec:sm-dcl}),  $m$ is set to $2$.
Furthermore, as shown in \cref{tab:mixed_scenes_sm_dcl,tab:start_epoch_sm_dcl}, we experimented with various numbers of mixed scenes and different mixing timings in SM-DCL, and adopted the parameters that yielded the highest accuracy.
The hyperparameters are simply found in a standard coarse-to-fine grid search or step-by-step tuning.
\begin{table*}[t]
  \begin{minipage}[tl]{0.27\textwidth}
  \caption{Accuracy degradation with studies on increasing number of mixed scenes on UCF101-24.}
    \centering
    \scalebox{0.7}{
    \begin{tabular}{c|ccc} \hline
    \# mixed scenes  & $1$ & $2$ & $4$ \\ \hline
    AP (\%) & $31.2$ & $34.1$ & $30.4$ \\ \hline
    \end{tabular}}
    \label{tab:mixed_scenes_sm_dcl}
    \caption{Ablation study on varying the starting epoch for applying SM-DCL using UCF101-24.}
    \centering
    \scalebox{0.7}{
        \begin{tabular}{c|ccc} \hline
            Starting epoch & $0$ &$20$ & $35$ \\ \hline
            AP (\%) & $27.3$ & $34.1$ & $33.2$ \\ \hline
        \end{tabular}
    }
    \label{tab:start_epoch_sm_dcl}
     \captionof{table}{Accuracy degradation with an increasing number of mixed scenes on UCF101-24.}
  \centering
  \scalebox{0.7}{
    \begin{tabular}{c|cc} \hline
    \# mixed scenes & 1 & 2 \\
    Average \# people & $1.2$ & $2.4$ \\ \hline
    AP (\%) & $31.2$ & $30.0$ \\ \hline
    \end{tabular}
  }\label{tab:incl_multi_act}
  \end{minipage}
  \hspace{0.001\textwidth}
  \begin{minipage}[tl]{0.33\textwidth}    
    \centering
    \includegraphics[width=1\textwidth]{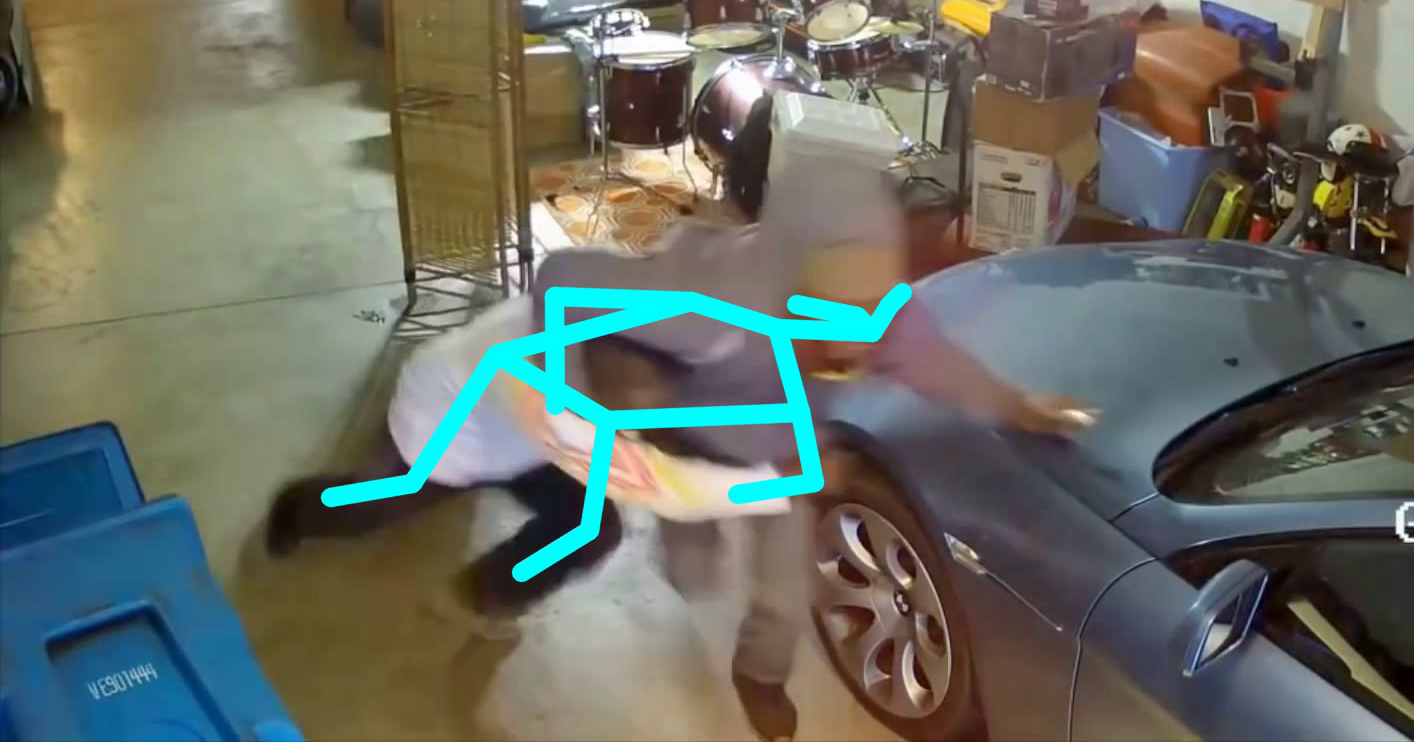}
    \captionof{figure}{Qualitative results for skeletons classified as violent behavior by the proposed method in RWF-2000.}
    \label{fig:quality_rwf}
    \captionof{table}{UCF101-24 average accuracy for the three action classes with the lowest similarity to those included in the pretraining dataset.}
    \centering
    \scalebox{0.6}{
        \begin{tabular}{c|c} \hline
            Action class & AP (\%) \\ \hline
            Selected three classes & $25.7$ \\
            All & $31.2$ \\ \hline
        \end{tabular}
    }
    \label{tab:low_sim}    
    \end{minipage}
    \hspace{0.001\textwidth}
  \begin{minipage}[tl]{0.38\textwidth}
    \caption{Ablation study on text encoders on UCF101-24.}
      \centering
      \scalebox{0.67}{
        \small
          \begin{tabular}{c|c} \hline
              Text encoder & AP (\%)  \\ \hline
              MiniLM~\cite{minilm} & $29.1$  \\
              CLIP ViT-L/14~\cite{pmlr-v139-radford21a} & $29.8$  \\
              MPNet~\cite{NEURIPS2020_c3a690be} & $31.2$  \\ \hline
          \end{tabular}
      }
      \label{tab:sentence_encoder}
    \caption{Ablation study on scaling up the pretraining dataset using UCF101-24.}
    \scalebox{0.67}{
        \begin{tabular}{c|ccc} \hline
            Ratio of scale & $0.25$ & $0.5$ & $1.0$  \\ \hline
            AP (\%) & $21.0$ & $23.1$ & $31.2$  \\ \hline
        \end{tabular}
    }
    \label{tab:scale_pretrain}
        \caption{Ablation study on text prompts using RWF-2000.}
    \centering
    \scalebox{0.67}{
    \begin{tabular}{c|cc} \hline
    Sentence & Acc. (\%) \\ \hline
    \multirow{3}{*}{\shortstack{\texttt{push or punch} \\ \texttt{related to } \\ \texttt{violent scene}}} & \multirow{3}{*}{$83.5$}\\
    & \\
    & \\ \hline
    \multirow{3}{*}{\shortstack{\texttt{push, kick, punch} \\ \texttt{or drag related}\\ \texttt{to violent scene}}} & \multirow{3}{*}{$82.0$}\\ 
    & \\
    & \\ \hline
    \multirow{3}{*}{\shortstack{\texttt{push, kick} \\ \texttt{or punch related}\\ \texttt{to violent scene}}} & \multirow{3}{*}{$84.0$} \\ 
    & \\
    & \\ \hline
    \end{tabular}
    \label{tab:CompText}
    }
    \end{minipage}
\end{table*}

\begin{table}[t]
     \caption{Ablation study on the mixing ratios of action classification loss and contrastive loss.}
    \vspace{-0.2em}
    \centering
    \scalebox{0.8}{
    \begin{tabular}{c|ccccc} \hline
            Loss weight & $0.0$ & $0.2$ & $0.4$ & $0.6$ & $0.8$ \\ \hline
            AP (\%) & $24.8$ & $30.4$ & $30.5$ & $31.2$ & $30.1$ \\ \hline
    \end{tabular}
    \label{tab:CompLossWeight}}
\end{table}

\begin{table}[t]
     \caption{Ablation study on pooling switching mechanism on UCF101-24.}
    \vspace{-0.2em}
    \scalebox{0.8}{
        \begin{tabular}{c|c} \hline
            Method & AP (\%) \\ \hline
            Ave. Pool.~\cite{ICASSP2019_Wang} & $30.6$ \\
            Attention~\cite{ICASSP2019_Wang} & $26.4$ \\ \hline
            Max Pool. (Ours) & $31.2$ \\ \hline
        \end{tabular}
    }
    \label{tab:pooling_switch}
    \vspace{4.5mm}
    \centering
    \caption{Ablation study on contrastive learning loss function using UCF101-24.}
    \vspace{-0.2em}
    \scalebox{0.8}{
        \begin{tabular}{c|c} \hline
            Method & AP (\%) \\ \hline
            L2 norm loss & $28.8$ \\ 
            Barlow Twins~\cite{Barlow_2021_ICML} & $29.9$ \\ \hline
            Ours & $31.2$ \\ \hline
        \end{tabular}
    }
    \label{tab:contrastive_loss}
\end{table}
\begin{figure*}[h]
  \begin{center}
      \includegraphics[width=1\linewidth]{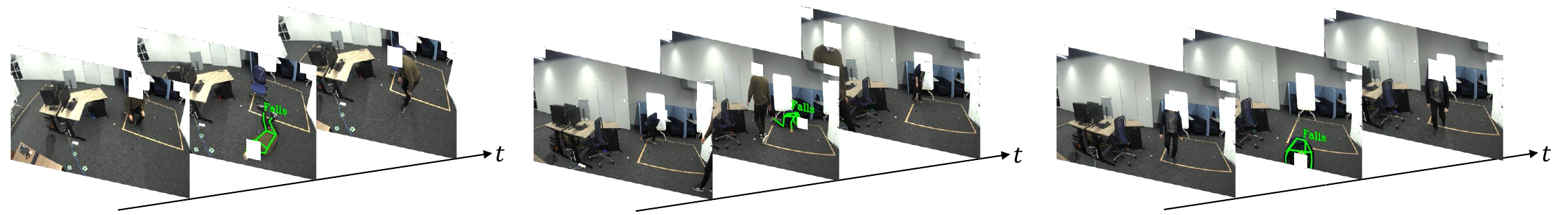}
      \caption{Qualitative zero-shot spatio-temporal action localization results of the proposed method on the FDD dataset. The input keypoints, including the fall behaviors, are visualized in the figure.}
      \label{fig:qualitative_resulst_fdd}
  \end{center}
\end{figure*}

\begin{table*}[tb]
  \caption{Summary of spatio-temporal action localization approaches.}
  \centering
  \scalebox{0.8}{
  \begin{tabular}{c|c|c|c|c} \hline
    \multirow{2}{*}{Method} & \multirow{2}{*}{Input information} & \multirow{2}{*}{\shortstack{\\ Learn target actions \\ in training (\xmark) or not (\cmark)}} & \multirow{2}{*}{\shortstack{\\ Supervision \\ at instance (\xmark) or video (\cmark) level}} & \multirow{2}{*}{\shortstack{\\ Multi-person \\ tracking free}} \\ 
    & & & & \\ \hline
    \cite{Singh_2018_ACCV,Singh_2017_ICCV,Li_2018_ECCV,Kalogeiton2017ICCV,Zhao_2019_CVPR} & RGB & \multirow{6}{*}{\xmark} &  & \xmark \\ 
    \cite{Duan_2023_ICCV} & Skeleton &  & \xmark & \xmark \\
    \cite{Pan_2021_CVPR,Feichtenhofer_2019_ICCV,tong2022videomae,Wang_2023_CVPR,Zhao_2022_CVPR} & RGB &  &  & \cmark \\ \cline{4-4}
    \cite{Anurag2020ECCV,Cheron2018Neurips,Victor2020CVIU} & RGB &  &  & \xmark \\ 
    \cite{Biswas_2021_ICCV} & RGB &  & \cmark & \cmark \\ 
    \cite{hfdt} & Skeleton &  &  & \xmark \\ \hline 
    \cite{Huang_2023_ICCV, bao2025exploiting, huang2025spatio} & RGB & \cmark & \xmark & \cmark \\  \hline
    Ours & Skeleton & \cmark & \cmark & \cmark \\ \hline
  \end{tabular}
  }
  \label{tab:summary_related_works}
\end{table*}
\subsection{Ablation Studies}

\noindent \textbf{Single-action assumption per video.}
During pretraining, the proposed method assumes that each video contains only one action.
However, many Kinetics-400 videos used for pretraining include scenes in which multiple people perform distinct actions (approximately $20$\%).
Moreover, we conducted an experiment during pretraining to assess the accuracy degradation of the proposed method when the skeletons from another randomly selected video are added as noise.
As shown in \cref{tab:incl_multi_act}, the accuracy degradation is only $1.2$\% points on the UCF101-24 dataset, indicating that the assumption is not essential for the proposed method.

\noindent \textbf{Generalization to outlier actions.}
Among the actions shown in \cref{tab:combined_results}, some outlier actions do not resemble those learned during pretraining.
Therefore, we analyzed the accuracy specifically for these outlier actions.
We compared the text feature representations of the action class names in the UCF101-24 dataset with those derived during pretraining.
\cref{tab:low_sim} shows the accuracy for the three outlier actions with the lowest similarity.
As the average accuracy for the outlier actions remains within a few percentage points of that for all action classes, we can conclude that the proposed method successfully detects outlier action classes.

\noindent \textbf{Analysis of SM-DCL}
In the SM-DCL method (\cref{sec:sm-dcl}), we set $m$, \ie, the number of mixed scenes, to $2$ based on experiments with different numbers of mixed scenes (\cref{tab:mixed_scenes_sm_dcl}) and various mixing timings (\cref{tab:start_epoch_sm_dcl}).
We then adopted the settings that achieved the highest accuracy.

\noindent \textbf{Analysis of Text Encoder Effectiveness}
\cref{tab:sentence_encoder} summarizes the differences in the performance of the proposed method on the UCF101-24 dataset using different text encoders.
Results show the effectiveness of the proposed method with different module combinations.


\noindent \textbf{Scaling up pretraining dataset}
We compared the accuracy of the proposed method on the UCF101-24 dataset using several subsets extracted at different ratios from the pretraining dataset (Kinetics-400).
\cref{tab:scale_pretrain} shows that the accuracy of the proposed method increases as the number of subsets used for pretraining gradually increases.

\noindent \textbf{Text prompts.}
\cref{tab:CompText} shows the violent recognition accuracy of the proposed method using four text prompts based on the RWF-2000 dataset.
Results show that the proposed method can localize unknown actions with some accuracy, regardless of the variations in text prompts.

\noindent \textbf{Mixing ratios of the action classification loss and the contrastive loss.}
\cref{tab:CompLossWeight} shows the spatio-temporal action localization accuracy of the proposed method using different mixing ratios of action classification and contrastive losses based on the UCF101-24 dataset.
Result shows that the proposed method that uses both loss terms achieves higher accuracy than when only the contrastive loss ($\alpha=0$) is employed, indicating the effectiveness of incorporating the action classification loss in the pretraining phase.

\noindent \textbf{Analysis of individual components.}
The proposed method is designed based on ablation studies that could not be included in the main paper due to space constraints. 
Although the details are presented in the final version (\cref{tab:pooling_switch,tab:contrastive_loss}), the pooling switching mechanism and contrastive learning loss function are adopted after comparing multiple modules.

\section{Human Pose Detectors}\label{sec:pose_detectors}
In this section, we provide detailed information about the pose detectors.

\noindent \textbf{PPNs.}
PPNs~\cite{Sekii2018ECCV} detect human skeletons in a {\it bottom-up} manner from an RGB image at high speed.
They comprise a ResNet-101 backbone~\cite{Wang2018Neurips} and trained using the MS-COCO dataset~\cite{Lin2014ECCV}.
The definition of the human skeleton is the same as OpenPose~\cite{8765346}.
As inputs to PPNs, we resize video frames to $640 \times 480$~$\text{px}^2$.

\noindent \textbf{HRNet.}
HRNet~\cite{Sun2019CVPR} is a {\it top-down} human pose detector.
It achieves superior accuracy, and the computational cost includes a human detector (Faster R-CNN~\cite{Ren2015Neurips}), which is expensive.
In the experiments on UCF101-24, we employ publicly available HRNet skeletons\footnote{\url{github.com/kennymckormick/pyskl}} by Haodong~\etal~\cite{Duan_2022_CVPR}.

\section{Qualitative Results on the FDD Dataset}
\cref{fig:qualitative_resulst_fdd} shows that the qualitative results of action localization obtained using the FDD dataset.
The proposed method localizes the falling actions of each person in each frame in a zero-shot manner.

\section{Related Work}
Herein, we briefly review related work in action recognition to provide context for our contributions.
Additionally, \cref{tab:summary_related_works} compares the features of the proposed method with related works, including those discussed in \cref{sec:related_work}.

\subsection{Action Recognition}
As mentioned in \cref{sec:intro}, three primary tasks have been extensively pursued in the field of action recognition: classification~\cite{Yan_2022_CVPR,Xing_2023_CVPR,hachiuma2023cvpr,Duan_2022_CVPR,Sijie2018AAAI,Piergiovanni_2023_CVPR,Su2020ECCV}, temporal localization~\cite{Shao_2023_ICCV,Miki_2020_WACV}, and spatio-temporal localization~\cite{Singh_2018_ACCV,Singh_2017_ICCV,Li_2018_ECCV,Kalogeiton2017ICCV,Zhao_2019_CVPR,Duan_2023_ICCV,Pan_2021_CVPR,Feichtenhofer_2019_ICCV,tong2022videomae,Wang_2023_CVPR,Zhao_2022_CVPR,Anurag2020ECCV,Cheron2018Neurips,Victor2020CVIU,Biswas_2021_ICCV,hfdt,Huang_2023_ICCV} of actions.
This study focuses on a skeleton-based spatio-temporal action localization task, wherein actions are recognized for each skeleton detected in every frame.
To extract spatio-temporal features in a data-driven manner from videos over the three tasks, appearance-based approaches employ 3D convolutional neural networks (CNNs) for spatio-temporal convolutions~\cite{Feichtenhofer_2019_ICCV,Pan_2021_CVPR} and transformers for capturing a global context of feature relationships across space and time~\cite{Yan_2022_CVPR,Piergiovanni_2023_CVPR,tong2022videomae,Wang_2023_CVPR,Zhao_2022_CVPR} given the recent advancement of DNN.
In contrast, skeleton-based approaches use graph convolutional networks~\cite{Sijie2018AAAI,Duan_2023_ICCV} and DNNs inspired by a point cloud deep learning paradigm~\cite{Su2020ECCV,hachiuma2023cvpr} to model the motion features from input skeleton sequences.
The proposed method exploits the skeleton-based approach, which is more resistant to changes in a person's appearance or background due to training~\cite{Weinzaepfel2021IJCV}.

\end{document}